# Approximation of Lorenz-Optimal Solutions in Multiobjective Markov Decision Processes

**Patrice Perny**
UPMC, LIP6
patrice.perny@lip6.fr

**Paul Weng**
UPMC, LIP6
paul.weng@lip6.fr

**Judy Goldsmith**
University of Kentucky
goldsmit@cs.uky.edu

**Josiah P. Hanna**
University of Kentucky
jpha226@g.uky.edu

## Abstract

This paper is devoted to fair optimization in Multiobjective Markov Decision Processes (MOMDPs). A MOMDP is an extension of the MDP model for planning under uncertainty while trying to optimize several reward functions simultaneously. This applies to multiagent problems when rewards define individual utility functions, or in multicriteria problems when rewards refer to different features. In this setting, we study the determination of policies leading to Lorenz-non-dominated tradeoffs. Lorenz dominance is a refinement of Pareto dominance that was introduced in Social Choice for the measurement of inequalities. In this paper, we introduce methods to efficiently approximate the sets of Lorenz-non-dominated solutions of infinite-horizon, discounted MOMDPs. The approximations are polynomial-sized subsets of those solutions.

## 1 INTRODUCTION

Planning under uncertainty is a central problem in developing intelligent autonomous systems. This problem is often represented by a Markov Decision Process (MDP) that provides a general formal framework for optimizing decisions in dynamic systems [2, 11]. Applications of MDPs occur in contexts such as robotics, automated control, economics, and manufacturing. The MDP model is characterized by a set of possible states, a set of possible actions enabling transitions from states to states, a reward function that gives the immediate reward generated by any admissible action $a$ is any state $s$, and a transition function that gives, for any state-action pair $(a, s)$, the resulting probability distribution over states. In such problems, the aim is to identify an optimal policy, i.e., a sequence of decision rules giving, at any stage of the process, and in any state, the action that must be selected so as to maximize the expected discounted reward over the long run.

However, there are various planning contexts in which the value of a policy must be assessed with respect to different point of views (individual utilities, criteria) and is not necessarily representable by a single reward function. This is the case in multiagent planning problems [3, 9] where every agent may have its own value system and its own reward function. This is also the case of multiobjective problems [1, 15, 4], for example path-planning problems under uncertainty when distance, travel time, and energy consumption are to be minimized simultaneously.

In such problems, resorting to $n$ distinct reward functions is natural, so as to express utilities of actions with respect to the different objectives. Hence, MDPs are generalized into MOMDPs (Multiobjective Markov Decision Processes) by extending the reward function to map a state-action pair to a reward vector which assigns a scalar reward for each objective. The value function will also be vector-valued, and the Bellman equation will continue to define the value of a policy in all states [27]. Note that a policy that maximizes on one objective will not necessarily do the same for another. Some policies will favor one objective, some another objective, and some will be balanced towards all objectives. Even when the reward functions could be aggregated linearly, keeping them separate enables a better control of tradeoffs and better recommendation possibilities. This explains the current interest for multiobjective (multicriteria or multiagent) extensions of Markov Decision Processes in the literature [15, 4, 3, 9, 16].

When several objectives must be optimized simultaneously, most of the studies on Markov Decision Processes concentrate on the determination of the entire set of Pareto optimal feasible tradeoffs, i.e., reward vectors (corresponding to feasible policies) that can-not be improved on one objective without being downgraded on another objective. However, when randomized policies are allowed, there are infinitely many such policies. Furthermore, when only deterministic policies are allowed, there are instances of MDPs in which the size of the Pareto set grows exponentially with the number of states, thus making its exact determination intractable.

In practical cases however, there is generally no need to determine the entire set of Pareto-optimal feasible tradeoffs, but only a reduced sample of solutions representative of the diversity of feasible tradeoffs. For this reason, some authors propose to work on the determination of a polynomially sized approximation of the Pareto set covering within a given threshold all feasible tradeoffs [19]. When richer preference information is available, an alternative approach consists in optimizing a scalarizing function measuring the value of any feasible reward vector [25]. In multicriteria optimization, the scalarizing function can be any preference aggregation function monotonically increasing with Pareto dominance or any measure of the distance to a given target in the space of criteria (reference point approach, [28]). In multiple agents problems, the scalarizing function can be any Social Welfare Function aggregating individual rewards.

In this paper we propose a third approach that consists in focusing the search on Lorenz-optimal tradeoffs, i.e., Pareto-optimal tradeoffs achieving a fair sharing of rewards among objectives. Lorenz dominance (L-dominance for short) is a partial preference order refining Pareto-dominance while including an idea of fairness in preferences. It is used for the measurement of inequalities in mathematical economics [24], for example to compare income distributions over a population. In our context, it can be used to compare reward vectors by inspecting how they distribute rewards over components. L-dominance is grounded on an axiomatic principle stating that any policy modification that induces a reward transfer reducing inequalities in the satisfaction of objectives will improve the solution. Within the Pareto-set, the subset of Lorenz-optimal solutions deserve special attention because it includes all tradeoffs of interest provided a balanced reward vector is sought. Moreover, the definition of Lorenz dominance does not require any specific preference information (neither weights nor target tradeoffs), beyond the fact that there is a preference for fair solutions.

The paper is organized as follows: in Section 2 we introduce basic concepts for MOMDPs, Pareto optimality and Lorenz optimality. Section 3 presents approximate optimality concepts for multiobjective problems and establishes preliminary results concerning the construction of minimal approximation of L-optimal tradeoffs. In Section 4, we describe a general method based on linear programming to approximate the set of L-optimal solutions in the case of $n$ objectives ($n \geq 2$) and a greedy algorithm to find approximation of minimal cardinality in the bi-objective case. Finally, in Section 5 we describe numerical tests on random instances of MDPs showing the efficiency of the proposed approaches.

## 2 BACKGROUND

### 2.1 MARKOV DECISION PROCESSES

A *Markov Decision Process (MDP)* is a tuple $\langle S, A, p, r \rangle$ where: $S$ is a finite set of states, $A$ is a finite set of actions, $p : S \times A \times S \to [0, 1]$ is a transition function giving, for each state and action the probability of reaching a next state, and $r : S \times A \to \mathbb{R}$ is a reward function giving the immediate reward for executing a given action in a given state [22].

Solving an MDP amounts to finding a *policy*, i.e., determining which action to choose in each state, which maximizes a performance measure. In this paper, we focus on the *expected discounted total reward* as the performance measure. A policy $\pi$ is called *deterministic* if it can be defined as a function from states to actions, i.e., $\pi : S \to A$. A policy $\pi$ is called *randomized* if for each state, it defines a probability distribution over actions, i.e., $\pi : S \times A \to [0, 1]$ where $\forall s, \sum_a \pi(s, a) = 1$. Note that a deterministic policy is a special case of randomized policy, i.e., $\forall s, \forall a, \pi(s, a) \in \{0, 1\}$. The expected discounted total reward for a randomized policy $\pi$ in a state $s$ can be obtained as a solution of the following equation:

$$V^\pi(s) = \sum_{a \in A} \pi(s, a)[r(s, a) + \gamma \sum_{s'} p(s, a, s')V^\pi(s')]. \quad (1)$$

Function $V^\pi : S \to \mathbb{R}$ is called the *value function* of $\pi$.

A policy whose value function is maximum in every state is an *optimal* policy. In (infinite horizon, discounted) MDPs, an optimal deterministic policy is known to exist. Such an optimal policy can be found using linear programming or dynamic programming techniques such as value iteration or policy iteration [22].

### 2.2 MULTIOBJECTIVE MDP

A *Multiobjective MDP* (MOMDP) is defined as an MDP with the reward function replaced by $r: S \times A \to \mathbb{R}^n$ where $n$ is the number of criteria, $r(s, a) = (r_1(s, a), \ldots, r_n(s, a))$ and $r_i(s, a)$ is the immediate reward for objective $i$. Now, a policy $\pi$ is valued by a value function $V^\pi : S \to \mathbb{R}^n$, which gives the expected discounted total reward vector in each state and can be computed with a vectorial version of (1) where additions and multiplications are componentwise.

Although MOMDPs can be used to solve some centralized planning problems involving multiple agents, it should not be confused with Multiagent MDPs (MMDPs) introduced in [3], which are models for coordinating agents having independent actions but a common reward function. In MOMDPs, actions are not necessarily "distributed" over agents and rewards are valued by vectors (one per agent) whereas in MMDPs, there is a single common objective and a consensus in the evaluation of states; moreover actions are distributed over agents.

To compare the value of policies in a given state $s$, the basic model adopted in most previous studies [8, 26, 27] is *Pareto dominance* (P-dominance for short). The *weak Pareto-dominance* is defined as follows: $\forall v, v' \in \mathbb{R}^n, v \succsim_P v' \Leftrightarrow \forall i = 1, \ldots, n, v_i \geq v'_i$ where $v = (v_1, \ldots, v_n)$ and $v' = (v'_1, \ldots, v'_n)$ and *Pareto-dominance* as: $v \succ_P v' \Leftrightarrow v \succsim_P v'$ and $\text{not}(v' \succsim_P v)$. For a set $X \subset \mathbb{R}^n$, a vector $v \in X$ is said to be *P-dominated* if there is another vector $v' \in X$ such that $v' \succ_P v$; vector $v$ is said to be *P-optimal* is there is no vector $v'$ such that $v' \succ_P v$. For a set $X \subset \mathbb{R}^n$, the set of *Pareto-optimal* vectors of $X$, called *Pareto set*, is $\text{PND}(X) = \{v \in X : \forall v' \in X, \text{ not } v' \succ_P v\}$.

In MOMDPs, for a given probability distribution $\mu_s$ over initial states, a policy $\pi$ is preferred to a policy $\pi'$ if $\sum_s \mu_s V^\pi(s) \succ_P \sum_s \mu_s V^{\pi'}(s)$. Standard methods for MDPs can be extended to solve MOMDPs by finding Pareto-optimal policies. We recall the linear programming approach [26].

$$(\mathcal{P}_0) \quad \begin{array}{l} \max z_i = \sum_{s \in S} \sum_{a \in A} r_i(s, a) x_{sa} \quad i = 1, \ldots, n \\ \sum_{a \in A} x_{sa} - \gamma \sum_{s' \in S} \sum_{a \in A} x_{s'a} p(s', a, s) = \mu_s \; \forall s \in S \\ x_{sa} \geq 0 \; \forall s \in S, \forall a \in A \end{array}$$

Recall that there is a one-to-one mapping between variables $(x_{sa})$ satisfying constraints of $\mathcal{P}_0$ and randomized policies $\pi$ (i.e., $\pi(s, a) = x_{sa} / \sum_a x_{sa}$) and $\sum_{s \in S} \sum_{a \in A} R_i(s, a) x(s, a) = \sum_s \mu_s V_i^\pi(s)$ for all $i = 1, \ldots, n$. More specifically, the constraints of $\mathcal{P}_0$ define a polytope whose extreme points are deterministic policies. For a deterministic policy, in every state $s$, $x_{sa}$ is non-null only for one action $a$. Thus, solving this multiobjective linear program amounts to optimizing the objective function $\sum_{s \in S} \mu_s V(s)$, called the *value vector* and interpreted as the expectation of a vector value function $V$ w.r.t. probability distribution $\mu_s$.

Following [7], one could add the following constraints to this linear program, obtaining then a mixed linear program with $0, 1$ variables, to restrict the search to deterministic policies only:

$$\begin{array}{ll} \sum_{a \in A} d_{sa} \leq 1 & \forall s \in S \\ (1 - \gamma) x_{sa} \leq d_{sa} & \forall s \in S, \forall a \in A \\ d_{sa} \in \{0, 1\} & \forall s \in S, \forall a \in A. \end{array} \quad (2)$$

As Pareto dominance is a partial relation, there generally exist many Pareto-optimal policies. In fact, in the worst case, it may happen that the number of Pareto-optimal value vectors corresponding to deterministic policies is exponential in the number of states as shown in the following example, adapted from [10].

**Example 1** *Let $N > 0$. Consider the following deterministic MOMDP represented in Figure 1. It has $N+1$ states. In each state, two actions (Up or Down) are possible except in the absorbing state $N$. The rewards are given next to the arcs representing the two actions. Here, we can take $\gamma = 1$ as state $N$ is absorbing. In this example, there are $2^{N+1}$ stationary deterministic policies. Stationary deterministic policies that only differ from one another on the choice of the action in the last state $N$ have the same value functions as the reward and the transition in those states for both actions are identical. In the initial state 0, the remaining policies induce $2^N$ different valuation vectors, of the form $(x, 2^N - 1 - x)$ for $x = 0, 1, \ldots, 2^N - 1$. Those different vectors are in fact all Pareto-optimal as they are on the line $x + y = 2^N - 1$.*

This example suggests that computing all Pareto-optimal solutions is not feasible in the general case. Moreover, deciding whether there exists a deterministic policy whose value vector P-dominates a given vector is known to be NP-hard [4, 23].

In this paper, we want to determine a subset of the Pareto set containing only policies that fairly distribute rewards among agents. The aim of generating well-balanced solutions has been tackled with scalararizing functions such as max-min [18, 12], augmented Tchebycheff norm [21] or WOWA of regrets [16]. However, each of these criteria focuses on a very specific idea of fairness and can only be justified when we have a very precise preferential information. A more cautious approach is to rely on Lorenz domi-nance that is a partial order, leaving room for various optimally fair solutions. Let us now introduce more precisely the notions of Lorenz dominance and Lorenz optimality.

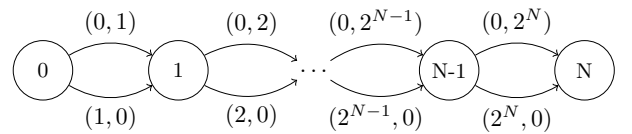

Figure 1: An instance where all deterministic policies have distinct Pareto-optimal value vectors

Lorenz dominance relies on a cautious idea of fairness, namely the *transfer principle*: Let $v \in \mathbb{R}_+^n$ such that $v_i > v_j$ for some $i, j$. Then for all $\varepsilon > 0$ such that $\varepsilon \leq v_i - v_j$, any vector of the form $v - \varepsilon e_i + \varepsilon e_j$ is preferred to $v$, where $e_i$ (resp. $e_j$) is the vector whose $i^{th}$ (resp. $j^{th}$) component equals 1, all others being 0.

This principle captures the idea of fairness as follows: If $v_i > v_j$ for some value vector $v \in \mathbb{R}_+^n$, slightly improving component $v_j$ to the detriment of $v_i$ while keeping the sum unchanged would produce a better distribution of rewards, and consequently a more suitable solution. Such transfers reducing inequalities are named *admissible* transfers also known as *Pigou-Dalton transfers* in Social Choice Theory. For example, value vector $(10, 10)$ should be preferred to $(14, 6)$ because there is an admissible transfer of size 4. Note that using a similar transfer of size greater than 8 would be counterproductive because it would increase inequalities in satisfaction. This explains why the transfers must have a size $\varepsilon \leq v_i - v_j$.

The transfer principle provides arguments to discriminate between vectors having the same average rewards. When combined with Pareto monotonicity (compatibility of preference with P-dominance), it becomes more powerful. For example, consider value vectors $(11, 11)$ and $(12, 9)$ respectively, we can remark that on the one hand, $(11, 11)$ is better than $(11, 10)$ due to Pareto dominance and $(11, 10)$ is better than $(12, 9)$ thanks to the Transfer Principle. Hence, we are able to conclude that $(11, 11)$ is better than $(12, 9)$ by transitivity. In order to better characterize those vectors that can be compared using improving sequences based on P-dominance and admissible transfers, we recall the definition of Lorenz vectors and Lorenz dominance (for more details see e.g. [14, 24]):

**Definition 1** *For all $v \in \mathbb{R}_+^n$, the* Lorenz Vector *associated to $v$ is the vector:*

$$L(v) = (v_{(1)}, v_{(1)} + v_{(2)}, \ldots, v_{(1)} + v_{(2)} + \ldots + v_{(n)})$$

*where $v_{(1)} \leq v_{(2)} \leq \ldots \leq v_{(n)}$ represents the components of $v$ sorted by increasing order. The $k^{th}$ component of $L(v)$ is $L_k(v) = \sum_{i=1}^{k} v_{(i)}$.*

**Definition 2** *Hence, the* Lorenz dominance *relation (L-dominance for short) on $\mathbb{R}_+^n$ is defined by:*

$$\forall v, v' \in \mathbb{R}_+^n, \ v \succsim_L v' \iff L(v) \succsim_P L(v')$$

*Its asymmetric part is defined by:*

$$v \succ_L v' \iff L(v) \succ_P L(v').$$

Within a set $X$, any element $v$ is said to be *L-dominated* when $v' \succ_L v$ for some $v'$ in $X$, and *L-optimal* when there is no $v'$ in $X$ such that $v' \succ_L v$. The set of L-optimal elements in $X$, called the *Lorenz set*, is denoted $\text{LND}(X)$. In order to establish the link between Lorenz dominance and preferences satisfying the combination of P-Monotonocity and the transfer principle we recall a result of [5]:

**Theorem 1** *For any pair of vectors $v, v' \in \mathbb{R}_+^n$, if $v \succ_P v'$, or if $v$ is obtained from $v'$ by a Pigou-Dalton transfer, then $v \succ_L v'$. Conversely, if $v \succ_L v'$, then there exists a sequence of admissible transfers and/or Pareto-improvements to transform $v'$ into $v$.*

This theorem establishes Lorenz dominance as the minimal transitive relation (with respect to inclusion) satisfying compatibility with P-dominance and the transfer principle. As a consequence, the subset of L-optimal value vectors appears as a very natural solution concept in fair optimization problems. A consequence of Theorem 1 is that $v \succ_P v'$ implies $v \succ_L v'$. Hence, L-dominance is a refinement of P-dominance and the set of L-optimal vectors is included in the set of P-optimal vectors.

The number of Lorenz-optimal tradeoffs is often significantly smaller than the number of Pareto-optimal tradeoffs. For instance in Example 1, while there is an exponential number of Pareto-optimal policies, with distinct value vectors, there are only two Lorenz-optimal policies with value vectors $(\lfloor \frac{2^N-1}{2} \rfloor, \lceil \frac{2^N-1}{2} \rceil)$ and $(\lceil \frac{2^N-1}{2} \rceil, \lfloor \frac{2^N-1}{2} \rfloor)$. Unfortunately, there exist instances where the number of Lorenz-optimal value vectors corresponding to deterministic policies is exponential in the number of states, as shown in the following example adapted from [20].

**Example 2** *Let $N > 0$. Consider the following deterministic MOMDP represented in Figure 2, which is an adaptation of Example 1. It has $N + 1$ states. In each state, two actions (Up or Down) are possible except in the absorbing state $N$. The rewards are given next to the arcs representing the two actions. Here, we can take $\gamma = 1$ as state $N$ is absorbing.*

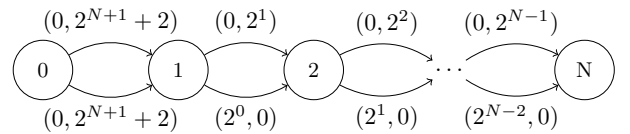

Figure 2: An instance where all deterministic policies have distinct Lorenz-optimal value vectors

*From state 1, all the possible value vectors are in the set $\{(x, 2(2^{N-1} - 1 - x)) | x = 0, 1, 2, \ldots, 2^{N-1} - 1\}$. Then, from the initial state $s_0$, the possible value vec-*

tors are in $\{(x, 3 \times 2^N - 2x) | x = 0, 1, 2, \ldots, 2^{N-1} - 1\}$. The set of Lorenz vectors is then $\{(x, 3 \times 2^N - x) | x = 0, 1, 2, \ldots, 2^{N-1} - 1\}$, implying that all the Lorenz vectors are Pareto-optimal.

This example shows that, although more discriminating than Pareto dominance, Lorenz dominance might leave many solutions incomparable. Therefore, on large instances, it may be infeasible to determine all Lorenz non-dominated solutions. Moreover, in deterministic MOMDPs, deciding whether there exists a policy whose value vector L-dominates a fixed vector is NP-hard [20].

## 3 APPROXIMATION OF PARETO AND LORENZ SETS

### 3.1 $\varepsilon$-COVERING OF NON-DOMINATED ELEMENTS

The examples provided at the end of Section 2 show that, even if we restrict ourselves to deterministic policies and two objectives, the set of P-optimal tradeoffs and the set of L-optimal tradeoffs may be very large. Their cardinality may grow exponentially with the number of states. Hence one cannot expect to find efficient algorithms to generate these sets exactly. This suggests that relaxing the notion of L-dominance (resp. P-dominance) to approximate the Lorenz set (resp. the Pareto set) with performance guarantees on the approximation would be a good alternative in practice. We first recall some definitions used to approximate dominance and optimality concepts in multiobjective optimization. We then investigate the construction of an approximation of the set of L-optimal tradeoffs. First, we consider the notion of $\varepsilon$-dominance defined as follows [19, 13]:

**Definition 3** For any $\varepsilon > 0$, the $\varepsilon$-dominance relation is defined on value vectors of $\mathbb{R}^n$ as follows:

$$x \succsim_P^\varepsilon y \Leftrightarrow [\forall i \in N, (1+\varepsilon)x_i \geq y_i].$$

Hence we can define the notion of approximation of the Pareto set as follows:

**Definition 4** For any $\varepsilon > 0$ and any set $X \subseteq \mathbb{R}^n$ of bounded value vectors, a subset $Y \subseteq X$ is said to be an $\varepsilon$-covering of $PND(X)$ if $\forall x \in PND(X), \exists y \in Y : y \succsim_P^\varepsilon x$.

For example, on the left part of Figure 3, the five black points form an $\varepsilon$-covering of the Pareto set. Indeed, dotted lines define 5 cones delimiting the areas where value vectors are $\varepsilon$-dominated by a black point. One can see that the union of these cones covers all feasible value vectors. Of course, a given set $X$ of feasible tradeoffs may include multiple $\varepsilon$-covering sets, set $X$ is itself an $\varepsilon$-covering of $X$. In practice, we are interested in finding an $\varepsilon$-covering the size of which is polynomially bounded.

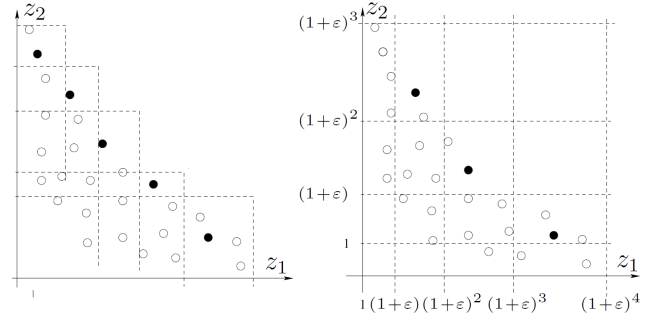

Figure 3: $\varepsilon$-coverings of the Pareto set

The strength of the $\varepsilon$-covering concept is derived from the following result of Papadimitriou and Yannakakis [19]: for any fixed number of criteria $n > 1$, for any finite $\epsilon > 0$ and any set $X$ of bounded value vectors such that $0 < x_i \leq K$ for all $i \in N$, there exists in $X$ an $\varepsilon$-covering of the Pareto set $PND(X)$ the size of which is polynomial in $\log K$ and $1/\epsilon$. The result can be simply explained as follows: to any reward vector $x \in \mathbb{Z}^n$, we can assign vector $\varphi(x)$ the components of which are $\varphi(x_i) = \lceil \frac{\log x_i}{\log(1+\varepsilon)} \rceil$. Due to the scaling and rounding operation, the number of different possible values for $\varphi$ is bounded on each axis by $\lceil \log K / \log(1+\varepsilon) \rceil$. Hence the cardinality of set $\varphi(X) = \{\varphi(x), x \in X\}$ is upper bounded by $\lceil \log K / \log(1+\varepsilon) \rceil^n$.

This can easily be illustrated using the right part of Figure 3 representing a logarithmic grid in the space of criteria. Any square of the grid represents a different class of value vectors having the same image through $\varphi$. Any vector belonging to a given square covers any other element of the square in terms of $\succsim_P^\varepsilon$. Hence, choosing one representative in each square, we cover the entire set $X$. If $n > 2$, squares become hypercubes. The size of the covering is bounded by the number of hypercubes in the hypergrid which is $\lceil \log K / \log(1+\varepsilon) \rceil^n$. The covering can easily be refined by keeping only the elements of $PND(\varphi(X))$ due to the following proposition:

**Proposition 1** $\forall x, y \in X, \varphi(x) \succsim_P \varphi(y) \Rightarrow x \succsim_P^\varepsilon y$.

Hence, remarking that for any fixed $x_1, \ldots, x_{n-1}$ there is no more than one Pareto-optimal element in vectors $\{\varphi(x_1, \ldots, x_{n-1}, z), z \in \mathbb{R}\}$ the $\varepsilon$-covering set will include at most $\lceil \log K / \log(1+\varepsilon) \rceil^{n-1}$ elements. In Example 1 where $n = 2$, if we consider the instance with 21 states, the Pareto set contains more than one million elements ($2^{20}$) whereas $\lceil \log 2^{20} / \log 1.1 \rceil = 146$ elements are sufficient to cover this set with a tolerance of 10% ($\varepsilon = 0.1$).

Similarly, we can define the notion of approximation of the Lorenz set as follows:

**Definition 5** *For any $\varepsilon > 0$ and any set $X \subseteq \mathbb{R}^n$ of bounded value vectors, a subset $Y \subseteq X$ is said to be an $\varepsilon$-covering of $LND(X)$ if $\forall x \in LND(X) \; \exists y \in Y$ : $y \succsim_L^\varepsilon x$, i.e. $L(y) \succsim_P^\varepsilon L(x)$.*

In other words, $Y$ is a $\varepsilon$-covering of $LND(X)$ if $L(Y) = \{L(y), y \in Y\}$ is a $\varepsilon$-covering $PND(L(X))$.

Hence, assuming that an algorithm $A$ generates an $\varepsilon$-covering of P-optimal elements in any set $X$, there are two indirect ways of constructing an $\varepsilon$-covering of $LND(X)$. The first way consists of computing $L(X) = \{L(x), x \in X\}$ and then calling $A$ to determine an $\varepsilon$-covering of $PND(L(X))$. This approach is easily implementable when the set of feasible tradeoffs $X$ is given explicitly. Unfortunately, in the case of MOMDPs as in many other optimization problems, the feasible set $X$ is only implicitly known. We show in Section 4 how this approach can be modified to overcome the problem in MOMDPs. A second way consists of first computing an $\varepsilon$-covering $Y$ of $PND(X)$ with $A$ and then determine $L(Y)$ and $PND(L(Y))$. This yields an $\varepsilon$-covering of $LND(X)$ as shown by:

**Proposition 2** *For any set $X$ of vectors, if $Y$ is an $\varepsilon$-covering of $PND(X)$, then $PND(L(Y))$ is an $\varepsilon$-covering of $LND(X)$.*

*Proof:* For any $x \in PND(X)$, there is a $y \in Y$ such that $y \succsim_P^\varepsilon x$. Hence $(1+\varepsilon)y \succsim_P x$ and $L((1+\varepsilon)y) \succsim_P L(x)$ by Theorem 1. Since $L((1+\varepsilon)y) = (1+\varepsilon)L(y)$ we obtain $(1+\varepsilon)L(y) \succsim_P L(x)$. Also, there is $z \in PND(L(Y))$ such that $L(z) \succsim_P L(y)$ and therefore $(1+\varepsilon)L(z) \succsim_P (1+\varepsilon)L(y)$. Hence by transitivity we get $(1+\varepsilon)L(z) \succsim_P L(x)$ and therefore $L(z) \succsim_P^\varepsilon L(x)$ □

The general result of Papadimitriou and Yannakakis [19] holds for MOMDPs, as shown by Chatterjee et al. [4]. For any MOMDP $\langle S, A, p, r \rangle$ with discount factor $\gamma \in (0,1)$, for all $\varepsilon > 0$, there exists an $\varepsilon$-covering of Pareto-optimal tradeoffs whose size is polynomial in $|S|$ (the number of states), $|\gamma|, |R|$ (an upper bound on rewards), and $1/\varepsilon$, and exponential in $n$. Moreover, there exists an algorithm to construct an $\varepsilon$-covering of the Pareto set in time polynomial in $|S|, |\gamma|, |R|$, and $1/\varepsilon$ and exponential in $n$. This algorithm is based on a systematic inspection of the squares of the grid given in Figure 1, using linear programming techniques. Hence when the number of criteria is fixed, Proposition 2 can be used to show the existence of a fully polynomial approximation scheme (fptas) for the set of L-optimal tradeoffs:

**Proposition 3** *For any fixed number of criteria $n > 1$, for any MOMDP $\langle S, A, p, r \rangle$ involving $n$ criteria and a discount factor $\gamma \in (0,1)$, for all $\varepsilon > 0$, there exists an $\varepsilon$-covering of Lorenz-optimal tradeoffs whose size is polynomial in $|S|, |\gamma|, |R|$, and $1/\varepsilon$. Moreover, there exists an algorithm to construct an $\varepsilon$-covering of Lorenz-optimal tradeoffs in time polynomial in $|S|, |\gamma|, |R|$, and $1/\varepsilon$.*

*Proof:* For any fixed $n > 1$, we know that an $\varepsilon$-covering of the Pareto set $Y$ of size polynomial in $|S|, |\gamma|, |R|$, and $1/\varepsilon$ can be computed in time polynomial in $|S|, |\gamma|, |R|$, and $1/\varepsilon$. Moreover, we have $PND(L(Y)) \subseteq L(Y)$ and $|L(Y)| \leq |Y|$, therefore $|PND(L(Y))|$ is polynomial in $|S|, |\gamma|, |R|$, and $1/\varepsilon$. Moreover, $L(Y)$ can be derived from $Y$ in polynomial time and then $PND(L(Y))$ is obtained from $L(Y)$ in polynomial time using pairwise comparisons. Proposition 2 concludes the proof since $PND(L(Y))$ is known to form an $\varepsilon$-covering of L-optimal solutions. □

Proposition 2 suggests a two-phase approach: first approximate the Pareto set and then derive an approximation of the Lorenz set. In the next section we investigate more direct methods to construct an approximation of the set of Lorenz-optimal tradeoffs.

## 4 DIRECT CONSTRUCTIONS OF $\varepsilon$-COVERING OF LORENZ SET

### 4.1 GENERAL PROCEDURE

We now present a direct procedure for constructing an $\varepsilon$-covering of the Lorenz set (of bounded size), without first approximating the Pareto set. This procedure relies on the observation made in the previous section: let $X$ be the set of feasible tradeoffs and $L(X)$ its image through the Lorenz transformation. Then $PND(L(X))$, the set of P-optimal vectors in $L(X)$ is an $\varepsilon$-covering of the Lorenz set.

Hence to any feasible tradeoff $x \in X$, we can assign a vector $\psi(x)$ where $\psi(x)_i = \lceil \frac{\log L_i(x)}{\log(1+\varepsilon)} \rceil$. Function $\psi$ defines a logarithmic hypergrid on $L(X)$ rather than on $X$. Any hypercube defined by $\psi$ in the hypergrid represents a class of value vectors that all have the same image through $\psi$. Any Lorenz vector $L(x)$ belonging to a given hypercube covers any other Lorenz vector $L(y)$ of the same hypercube in terms of $\succsim_P^\varepsilon$. Hence, the original vectors $x, y$ are such that $x \succsim_L^\varepsilon y$. Moreover the following property holds:

**Proposition 4** $\forall x, y \in X, \psi(x) \succsim_P \psi(y) \Rightarrow x \succsim_L^\varepsilon y$.

Thus, we can use P-optimal $\psi$ vectors to construct an $\varepsilon$-covering of the Lorenz set. Besides, due to the scaling and rounding operations, the number of different possible values for $\psi$ is bounded on the $i^{th}$ axis by $\lceil \log iK / \log(1+\varepsilon) \rceil$, where $K$ is an upper bound such that $0 < x_i \leq K$. Hence the cardinality of set $\psi(X) = \{\psi(x), x \in X\}$ is upper bounded by $\Pi_{i=1}^{n} \lceil \log(iK)/\log(1+\varepsilon) \rceil$. Moreover, since $L_i(x) \leq L_{i+1}(x)$ we have $\psi(x)_i \leq \psi(x)_{i+1}$ for all $i = 1, \ldots, n$. Therefore, when a $\psi(x)$ does not meet this constraint the corresponding hypercube is necessarily empty and does not need to be inspected. Thus the number of hypercubes that must be inspected is at most $\Pi_{i=1}^{n} \lceil \log(iK)/\log(1+\varepsilon) \rceil / n! \leq \Pi_{i=1}^{n} \lceil i \log K / \log(1+\varepsilon) \rceil / n! \leq \Pi_{i=1}^{n} i \lceil \log K / \log(1+\varepsilon) \rceil / n! = \lceil \log K / \log(1+\varepsilon) \rceil^n$. Hence, by choosing one representative in each of these hypercubes, we cover the entire set $L(X)$. The size of the covering is therefore bounded by $\Pi_{i=1}^{n} \lceil \log(iK)/\log(1+\varepsilon) \rceil / n!$, which is smaller than $\lceil \log K / \log(1+\varepsilon) \rceil^n$. Let us consider the following example:

**Example 3** *If $K = 10000$, $n = 3$ and $\varepsilon = 0.1$, the grid scanned in the Lorenz space includes $\Pi_{i=1}^{n} \lceil \log(iK)/\log(1+\varepsilon) \rceil / n! = 186,935$ hypercubes whereas the grid for the Pareto set includes $\lceil \log K / \log(1+\varepsilon) \rceil^n = 941,192$ hypercubes (5 times more).*

Thus, this approach is expected to be faster than the two-phase method presented in the previous section. This is confirmed by tests provided in Section 5. Moreover the resulting covering set can be reduced in polynomial time so as to keep only the elements of $\text{PND}(\psi(X))$. If any hypercube can be inspected in polynomial time, this direct approach based on the grid defined in the Lorenz space provides a fptas for the set of L-optimal value vectors in MOMDPs. Let us show now how hypercubes can be inspected using linear programming.

Let $z$ be the set of feasible value vectors $(z_1, \ldots, z_n)$ defined by $\mathcal{P}_0$ introduced in Section 2. We consider the following optimization problem designed to test whether there exists a feasible $z$ whose Lorenz vector $L(z)$ P-dominates a given reference vector $\eta \in \mathbb{R}^n$ representing the lower corner of an hypercube in the Lorenz space.

$$(\mathcal{P}_\eta) \quad \begin{array}{l} \max L_n(z) \\ L_k(z) \geq \eta_k, \quad k = 1, \ldots, n-1, \\ z \in Z. \end{array}$$

The objective of this optimization program is linear in variables $z_i$ since $L_n(z) = \sum_{i=1}^{n} z_i$. However, none of the constraints is linear since $L_k(z)$ is the sum of the $k$ greatest components of $z$ which requires sorting the components for every $z$. Fortunately, for any fixed $z$, the $k^{th}$ Lorenz component $L_k(z)$ can be defined as the solution of the following linear program [17]:

$$(\mathcal{P}_{L_k}) \quad \begin{array}{l} \min \sum_{i=1}^{n} a_{ik} z_i \\ \left\{ \begin{array}{l} \sum_{i=1}^{n} a_{ik} = k \\ a_{ik} \leq 1 \quad i = 1 \ldots n. \\ a_{ik} \geq 0 \quad i = 1 \ldots n. \end{array} \right. \end{array}$$

This does not directly linearize the constraints of $\mathcal{P}_\eta$ because $\mathcal{P}_{L_k}$ is a minimization problem and consequently $\sum_{i=1}^{n} a_{ik} z_i \geq \eta_k$ does not imply that $L_k(z) \geq \eta_k$. Fortunately, by duality theorem, $L_k(x)$ is also the optimal value of the dual problem of $\mathcal{P}_{L_k}$:

$$(\mathcal{D}_{L_k}) \quad \begin{array}{l} \max \quad kt_k - \sum_{i=1}^{n} b_{ik} \\ \left\{ \begin{array}{l} t_k - b_{ik} \leq z_i \quad i = 1 \ldots n \\ b_{ik} \geq 0 \quad i = 1 \ldots n. \end{array} \right. \end{array}$$

Since $\mathcal{D}_{L_k}$ is a maximization problem, imposing constraint $kr_k - \sum_{i=1}^{n} b_{ik} \geq \eta_k$ together with the constraints of $\mathcal{D}_{L_k}$ implies that $L_k(z) \geq \eta_k$. Hence we obtain the following linear reformulation of $\mathcal{P}_\eta$:

$$(LP_\eta) \quad \begin{array}{l} \max \sum_{k=1}^{n} z_k \\ \left\{ \begin{array}{l} kt_k - \sum_{i=1}^{n} b_{ik} \geq \eta_k, \; k = 1 \ldots n-1 \\ t_k - b_{ik} \leq z_i, \; i, k = 1 \ldots n \\ z \in Z. \\ b_{ik} \geq 0 \; i, k = 1 \ldots n. \end{array} \right. \end{array}$$

Finally, if $Z$ is the set of feasible value vectors of program $\mathcal{P}_0$ (see Section 2), then for any $p = (p_1, \ldots, p_n) \in \mathbb{N}^n$, one can test whether there exists a randomized policy, the Lorenz vector of which P-dominates vector: $\eta_\varepsilon^p = ((1+\varepsilon)^{p_1}, \ldots, (1+\varepsilon)^{p_n})$ by solving program $LP'_\eta$ below with $\eta = \eta_\varepsilon^p$ and checking that the objective at optimum is greater or equal to $(1+\varepsilon)^{p_n}$.

$$(LP'_\eta) \quad \begin{array}{l} \max \sum_{k=1}^{n} z_k \\ \left\{ \begin{array}{l} z_k = \sum_{s \in S} \sum_{a \in A} r_k(s,a) x_{sa}, \; k = 1 \ldots n \\ kt_k - \sum_{i=1}^{n} b_{ik} \geq \eta_k, \; k = 1 \ldots n-1 \\ t_k - b_{ik} \leq z_i, \; i, k = 1 \ldots n \\ b_{ik} \geq 0 \; i, k = 1 \ldots n \\ x_{sa} \geq 0 \; \forall s \in S, \forall a \in A \end{array} \right. \end{array}$$

Hence the whole logarithmic hypergrid in the Lorenz space can be entirely inspected using a polynomial number of calls to $LP'_{\eta_\varepsilon^p}$. One needs at most one call per integer valued vector $p \in \mathbb{N}^{n-1}$ such that $p_i \leq \lceil i \log K / \log(1+\varepsilon) \rceil$ and $p_1 \leq p_2 \leq \ldots \leq p_{n-1}$. These vectors $p$ are enumerated in lexicographic order.

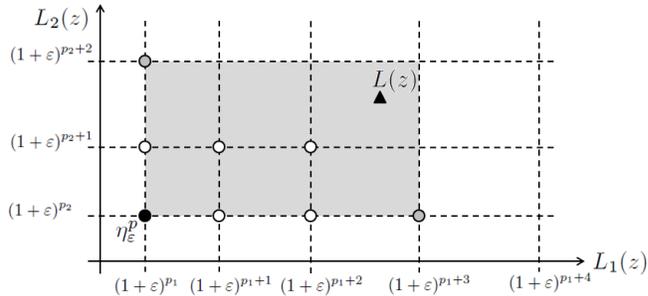

Figure 4: Grid in the Lorenz space

This systematic inspection can be significantly sped up due to the following observation illustrated in Figure 4: let $L(z)$ be the optimal Lorenz vector obtained by solving $LP'_{\eta^p_\varepsilon}$ (represented by a triangle in Figure 4). Then none of the hypercubes corresponding to an $\eta$ vector such that $L(z) \succsim_P \eta \succsim_P \eta^p_\varepsilon$ (white points in Figure 4) needs to be inspected because vectors falling in this area (colored in grey on Figure 4) are $\varepsilon$-dominated by $L(z)$. Hence calls to $LP'_\eta$ for such $\eta$ can be skipped to go directly to the next non-dominated $\eta$ vectors in the grid (grey points in Figure 4).

This procedure provides an $\varepsilon$-covering of L-optimal randomized pure policies. Whenever we want to restrict the search to deterministic policies, a similar procedure applies, we just need to add constraints given in Equation (2) as explained in Section 2. In this case, program $LP'_{\eta^p_\varepsilon}$ becomes a mix-integer linear program.

### 4.2 MINIMAL $\varepsilon$-COVERINGS: THE BIOJECTIVE CASE

The grid used to partition the entire space in the above procedure enables to avoid many unnecessary redundancies in the construction of a covering set because we keep at most one feasible policy in each hypercube. However, this procedure does not ensure that a covering of minimal cardinality will be found. In this subsection, we propose a greedy approach to generate an $\varepsilon$-approximation of minimal cardinality for the Lorenz set (and the Pareto set). The principle of this approach relies on a general scheme proposed in [6] for finding a minimal covering of the Pareto set in general biobjective optimization problems. Considering two objective functions $z_1$ and $z_2$, the construction consists in solving a sequence of optimization problems alternating two complementary subproblems:

**Restrict-1($\alpha_1$).** For any given value $\alpha_1$, we want to maximize $z_2$ subject to the constraint $z_1 \geq \alpha_1$. The procedure returns the optimal value vector or answers *no* when no such solution exists.

**Restrict-2($\alpha_2$).** For any given value $\alpha_2$, we want to maximize $z_1$ subject to the constraint $z_2 \geq \alpha_2$. The procedure returns the optimal value vector or answers *no* when no such solution exists.

Hence, the greedy construction of an $\varepsilon$-covering starts with the initial call $v_0 =$Restrict-2(0). Then we compute the following alternated sequences for $n \geq 1$:

$$u_n = \text{Restrict-1}(v_{n-1}/(1+\varepsilon))$$
$$v_{n+1} = \text{Restrict-2}((1+\varepsilon)u_n).$$

We let $n$ increase until the feasible domain of Restrict becomes empty. Point $v_0$ optimizes objective 1 but does not enter into the covering set. Instead we use $u_1$ which is, by construction, more "central" while still covering $v_0$. Then, we obtain $v_1$ as the rightmost Pareto-optimal point on the $z_1$ axis that is not covered by $u_1$. Like $v_0$, $v_1$ does not enter into the covering. Instead we include $u_2$ that improves $z_2$ while covering $v_1$ and so on. The resulting set $\{u_1, \ldots, u_q\}$ provides an $\varepsilon$-covering set of minimal cardinality. This procedure makes only $2q$ calls to Restrict. Further details on this greedy approach and its optimality for general biobjective problems are given in [6].

The specification of procedures Restrict-1 and Restrict-2 to construct an $\varepsilon$-covering of the Pareto set of minimal cardinality in biobjective MDPs is straightforward from $\mathcal{P}_0$. For constructing an $\varepsilon$-covering of minimal cardinality for the *Lorenz set* in biobjective MDPs we solve Restrict-i($\alpha_i$) using program $LP'_i(\alpha_i)$, for i=1,2, where $LP'_i(\alpha_i)$ is a convenient adaptation of $LP'_\eta$ defined as follows:

$$LP'_i(\alpha_i) \begin{cases} \max\ z_{3-i} \\ z_k = \sum_{s \in S}\sum_{a \in A} r_k(s,a)x_{sa},\ k=1,2 \\ it_i - b_{1i} - b_{2i} \geq \alpha_i, \\ t_k - b_{jk} \leq z_j,\ j,k=1,2 \\ b_{ik} \geq 0\ \ i,k=1,2 \\ x_{sa} \geq 0\ \ \forall s \in S, \forall a \in A. \end{cases}$$

Hence, Restrict-i($\alpha_i$) can be solved in polynomial time for randomized policies. Whenever we want to restrict the search to deterministic policies, we just have to add constraints given in Equation (2). In that case program $LP'_i$ is a MIP which cannot be expected to be solved in polynomial time. It is still easily solvable by current solvers, as is shown in the next section.

## 5 EXPERIMENTAL RESULTS

We tested the different methods presented in this paper on random instances of MOMDPs. The rewards on each objective were randomly drawn from $\{0, 1, \ldots, 99\}$. All the experiments were run on standard PCs with 8Gb of memory and an IntelCore 2 Duo 3.33GHz GHz processor. All LPs were solved using Gurobi 5.0. All the experimental results are averaged over 10 runs with discount factor $\gamma$ set to 0.9.

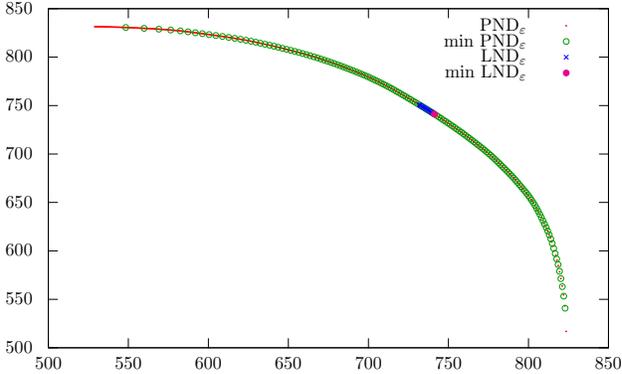

Figure 5: $\varepsilon$ approximation of Pareto and Lorenz sets

| $\varepsilon$ | 0.05 | 0.1 | 0.15 | 0.2 |
|---|---|---|---|---|
| L(PND$_\varepsilon$) | 265.7 | 169.4 | 126.7 | 101.7 |
| LND$_\varepsilon$ | 5.4 | 4.8 | 4.4 | 4.2 |

Table 1: Computation times of $\varepsilon$-covers

| $\varepsilon$ | 0 | 0.05 | 0.1 | 0.15 | 0.2 |
|---|---|---|---|---|---|
| PND$_\varepsilon$ | $2^{30}$ | 361 | 194 | 135 | 104 |
| L(PND$_\varepsilon$) | $2^{30}$ | 16 | 8 | 6 | 4 |
| min PND$_\varepsilon$ | $2^{30}$ | 15 | 8 | 5 | 4 |
| L(min PND$_\varepsilon$) | $2^{30}$ | 9 | 5 | 3 | 3 |
| LND$_\varepsilon$ | $2^{30}$ | 17 | 9 | 6 | 5 |
| min LND$_\varepsilon$ | $2^{30}$ | 4 | 2 | 2 | 1 |

Table 2: Sizes of $\varepsilon$-covers for the graph in Example 2

First, we illustrate how the size of $\varepsilon$-covers can be reduced using the greedy approach both for Pareto and Lorenz. In this first series of experiments, all the instances are biojective MDPs. We set the number of states to 200 and the number of actions to 5. Parameter $\varepsilon$ was set to 0.01. Figure 5 shows the value vectors in the objective space. PND$_\varepsilon$ (resp. LND$_\varepsilon$) is $\varepsilon$-cover of PND (resp. LND), min PND$_\varepsilon$ and min PND$_\varepsilon$ are the minimal $\varepsilon$-cover sets.

In the second series of experiments, we present the computation times (expressed in seconds) for computing the different $\varepsilon$-covers. In these experiments, the number of states is set to 50, the number of actions to 5 and the number of objectives to 3, and $\varepsilon \in \{0.05, 0.1, 0.15, 0.2\}$. The results are presented in Table 1 and shows that the direct approach is the most efficient. The longer computation time for the indirect method is mainly due to the determination of the approximate Pareto set.

To show the effectiveness of our approach, we computed the $\varepsilon$-covers for the graph presented in Example 2 with $N = 30$. The size of the $\varepsilon$-covers are given in Table 2.

## 6 CONCLUSION

We have proposed, compared and tested several efficient procedures for approximating (with performance guarantee) the set of Lorenz-optimal elements in MOMDPs. For randomized policies, the procedures presented are fully polynomial approximation schemes. Moreover for the bi-objective case, we presented a greedy approach which constructs, in polynomial time, for any given $\varepsilon > 0$, an $\varepsilon$-approximation of minimal cardinality of the set of Lorenz optimal tradeoffs. A similar approach works also for the Pareto set. We have shown how to modify this approach to determine covering sets using only deterministic policies. This has a computational cost since we have to solve MIPs instead of LPs. However, the numerical tests performed show that the approach remains efficient for deterministic policies on reasonably large instances. Moreover, the direct approximation of Lorenz-optimal elements enable to address larger problems than those requiring prior approximation of the Pareto set. Note that beside the `restrict` procedure, the approach is quite generic and could probably easily be adapted to other multiobjective problems.

These tools provide useful information for selecting optimal actions and policies in dynamic systems. By playing with threshold $\varepsilon$ we can increase or decrease on demand the size of our sample of solutions and provide more or less contrasted tradeoffs to cover the Pareto set or the Lorenz set. This approach could also be used to approximate $f$-optimal tradeoffs for any scalarizing function $f$ monotonic with respect to Pareto or Lorenz dominance.

Beside the application to aggregation functions, another research direction would be to look for procedures to construct minimal covering sets for MOMDPs involving more than two objectives. To the best of our knowledge this remains an open problem.

## Acknowledgments

Funded by the French National Research Agency under grant ANR-09-BLAN-0361; The Duncan E. Clarke Memorial Innovation Award, and NSF grants NSF EF-0850237 and CCF-1049360.